\documentclass[conference]{IEEEtran}
\usepackage{algorithm}
\usepackage{algorithmicx}
\usepackage{algpseudocode}
\usepackage{amsmath}
\usepackage{amsfonts}
\usepackage[hyphens]{url}
\usepackage{latexsym}
\usepackage{mathtools}
\usepackage{times}
\usepackage{xspace}

\usepackage{url}

\newcommand{\citet}[1]{\cite{#1}}

\newcommand\ModelI{BiLSTM-Softmax\xspace}
\newcommand\ModelII{BiLSTM-KB\xspace}
\newcommand\ModelIII{BiLSTM-Binary\xspace}
\newcommand\ModelIV{FastText-Softmax\xspace}

\begin{document}

\title{Deep Learning Approaches for Question Answering on Knowledge Bases: an evaluation of architectural design choices}

\author{\IEEEauthorblockN{Sherzod Hakimov}
\IEEEauthorblockA{Semantic Computing Group\\
CITEC, Bielefeld University}
\and
\IEEEauthorblockN{Soufian Jebbara}
\IEEEauthorblockA{Semantic Computing Group\\
CITEC, Bielefeld University}
\and
\IEEEauthorblockN{Philipp Cimiano}
\IEEEauthorblockA{Semantic Computing Group\\
CITEC, Bielefeld University}
}

\maketitle
\begin{abstract}
The task of answering natural language questions over knowledge bases has received wide attention in recent years. Various deep learning architectures have been proposed for this task. However, architectural design choices are typically not systematically compared nor evaluated under the same conditions. In this paper, we contribute to a better understanding of the impact of architectural design choices by evaluating four different architectures under the same conditions. We address the task of answering simple questions, consisting in predicting the subject and predicate of a triple given a question. In order to provide a fair comparison of different architectures, we evaluate them under the same strategy for inferring the subject, and compare different architectures for inferring the predicate. The architecture for inferring the subject is based on a standard LSTM model trained to recognize the span of the subject in the question and on a linking component that links the subject span to an entity in the knowledge base. The architectures for predicate inference are based on i) a standard softmax classifier ranging over all predicates as output, ii) a model that predicts a low-dimensional encoding of the property given entity representation and question, iii) a model that learns to score a pair of subject and predicate given the question as well as iv) a model based on the well-known FastText model. The comparison of architectures shows that FastText provides better results than other architectures.

\end{abstract}

\IEEEpeerreviewmaketitle

\section{Introduction}
The task of Question Answering (QA) has received increasing attention in the last few years. Most research has concentrated on the task of answering factoid questions such as \emph{Who wrote Mildred Pierced?}, yielding the answer \emph{Stuart Kaminsky}.
Typically, such answers are extracted from a knowledge base (KB).
A frequently used dataset in this context is the SimpleQuestions \cite{simpleQuestions} dataset, which consists of \emph{simple} questions that can be answered with a single fact from the Freebase KB. For instance, the question above can be answered using the following triple from Freebase:

\begin{small}
\begin{verbatim}
Subject:   m.04t1ftb (mildred_pierced)
Predicate: book.written_work.author
Object:    m.03nx4yz (stuart_kaminsky)
\end{verbatim}
\end{small}

The system needs to identify the relevant entity (subject), i.e. \emph{mildred\_pierced} in the example question, and infer the appropriate predicate, i.e. \emph{book.written\_work.author}.
In the case of SimpleQuestions, all questions involve a single triple, with the answer being the corresponding object. Thus, the task involves essentially predicting the subject and predicate of a triple.

Many different architectures have been proposed for this task, in particular many deep learning architectures. However, a systematic comparison of different architectural choices has not been provided so far. In particular, different property predicting systems have used different approaches to identifying the entity, so that they are not directly comparable.

Using a common model for entity prediction based on an NER architecture, we consider four different architectures for the predicate prediction task:

\begin{itemize}
\item \textbf{\ModelI}: this architecture uses a standard BiLSTM softmax classifier to predict the property in a question where the output ranges over all properties seen during training.
\item \textbf{\ModelII}: instead of using softmax layer output, this model predicts a low-dimensional representation of predicates that match to the closest predicate representation in pre-trained KB embeddings; the closest property is found using cosine similarity.
\item \textbf{\ModelIII}: this architecture outputs a binary decision on whether a pair of subject and predicate matches for the given question $q$ (true or false).
\item \textbf{\ModelIV}: this architecture uses FastText\footnote{\url{https://github.com/facebookresearch/fastText}} as a classifier to predict the property.

\end{itemize}

As a main contributions in this paper are:

\begin{itemize}
    \item Most systems do not report the performance of their individual components but just the overall score. This makes it hard to compare them on the sub-task level (entity linking, predicate classification, answer selection). We provided evaluations for all components in isolation under the same conditions.
    \item We compare different architectural choices to evaluate the performance of predicting the predicate for a given question answering task.
    \item We emphasize the importance of entity linking component and show how it affects overall performance on question answering task.
\end{itemize}

The paper is structured as follows: the next Section \ref{methods} describes our NER-based system for predicting the entity as well as the four architectures are described. Section \ref{evaluation} presents the results of our evaluation along with error analysis and discussion. Before concluding, we discuss the related work.

\section{Methods}
\label{methods}

\begin{figure}[ht]	
\centering
  \includegraphics[width=0.50\textwidth]{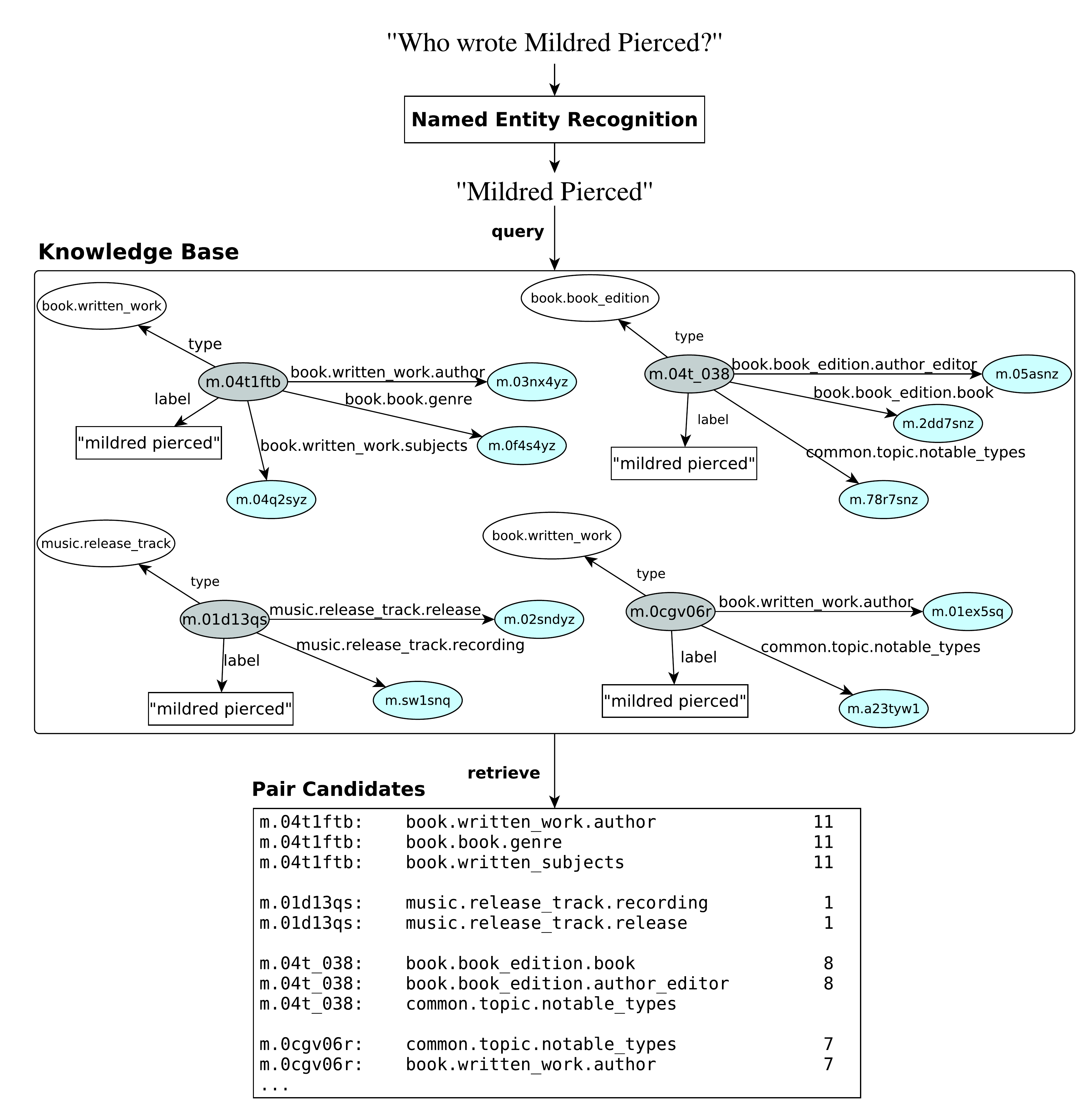}
	\caption{Visualization of a candidate pair generation process where named entities are queried on surface forms from the knowledge base}
	\label{neural:candidate}
\end{figure}

The task of answering simple questions requires identifying the correct entity and the predicate in the question. 
In this section, we describe in detail the model for identifying the span of the entity and candidates. Then, we describe four architectures for predicate prediction that build on this common entity prediction model. All four architectures rely on a candidate retrieval step that extracts candidate pairs of subject and predicate and then score pairs of subject/predicate to predict a query consisting of a single subject and predicate. 
The process is shown in Figure~\ref{neural:candidate}. In order to retrieve entity candidates we rely on an inverted index the construction of which we detail in the section below.

\subsection{Inverted Index Construction for Entity Retrieval}\label{subsec:neural-index}
We extract all entity mentions from Freebase using \emph{type.object.name} and \emph{common.topic.alias} predicates. During the extraction process, we also counted how often a surface form occurs together with an entity. As a result, we generated a surface form index for each subject with an associated frequency value. Additionally, we merged a surface form index created for DBpedia entities using \emph{owl:sameAs} links. Hakimov et al. \citet{hakimov2016A} provides such an index of surface forms. We converted the DBpedia URIs into Freebase MIDs using the links provided by the DBpedia release of 2014\footnote{\url{http://oldwiki.dbpedia.org/Downloads2014\#links-to-freebase}}. The converted index was merged with the index data extracted from Freebase. We aggregated the frequency values if the same surface form and Freebase URI (MID) existed in both indexes.

A sample from this index is given below. All surface forms in the index are normalized; they are converted into lowercase, punctuation as well as non-alpha-numeric characters are removed, etc.

\begin{table}[ht]
    \centering
    \begin{tabular}{c|c|c}
         Surface Form & URI & Frequency\\
         \hline
         mildred pierced & m.04t1ftb & 11\\
         mildred pierced & m.04t\_038 & 8\\
         mildred pierced & m.0cgv06r & 7\\
         \hline
    \end{tabular}
    \label{tab:my_label}
\end{table}

\subsection{Named Entity Recognition}\label{subsec:neural-ner}

We trained a Named Entity Recognizer (NER) system similar to the one proposed by Chiu and Nichols \cite{nerSimpleQA} using weak supervision, for which Raj \cite{nerSimpleQA-Code} provided the implementation. Since the dataset requires a single subject we adapted the NER to identify a single entity span. 

The original approach is tailored towards identifying common named entity (NE) types: LOCATION, PERSON, ORGANISATION, MISCELLANEOUS. Our goal is extract the single named entity span without doing any distinction between those types. We use an \emph{IO} tagging scheme to mark tokens inside (I) and outside (O) of the single named entity of interest.

\begin{figure}[ht]	
\centering
  \includegraphics[width=0.45\textwidth]{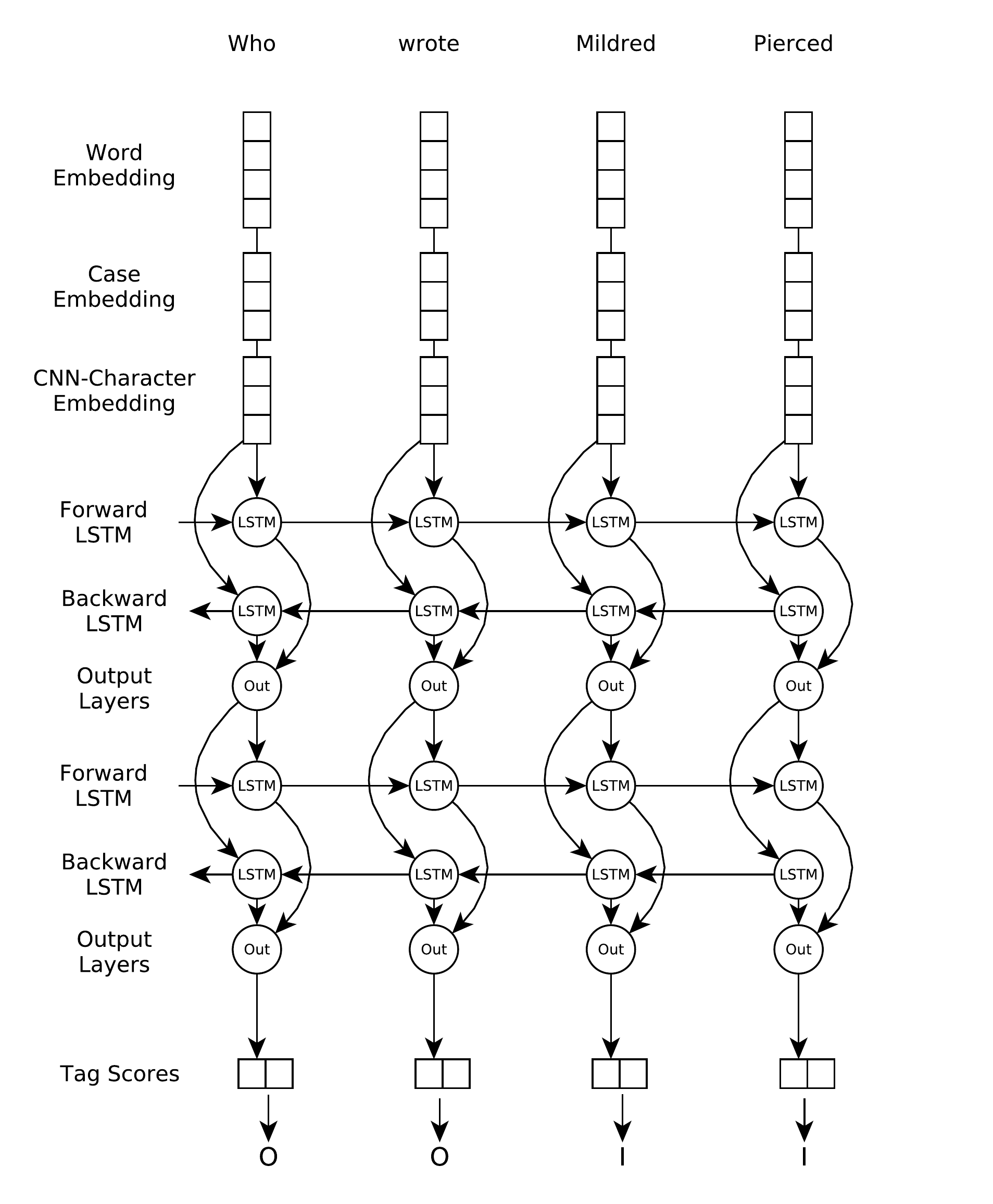}
	\caption{Named Entity Recognition using Bidirectional LSTM}
	\label{neural:ner}
\end{figure}

We merge the consecutive tokens that have \emph{I} as an output. This process is illustrated in Figure~\ref{neural:ner}. The predicted output shows that tokens \emph{Mildred} and \emph{Pierced} get assigned the output \emph{I} while other tokens get \emph{O} as an assigned label.

The architecture is based on Bidirectional LSTMs (BiLSTM) \cite{graves2013speech}. It is composed of two LSTM \cite{lstm} layers. The model uses words and characters as features along with case of words (lowercase, uppercase). These features are concatenated and fed into a neural network. 

\begin{figure}[ht]	
\centering
  \includegraphics[width=0.45\textwidth]{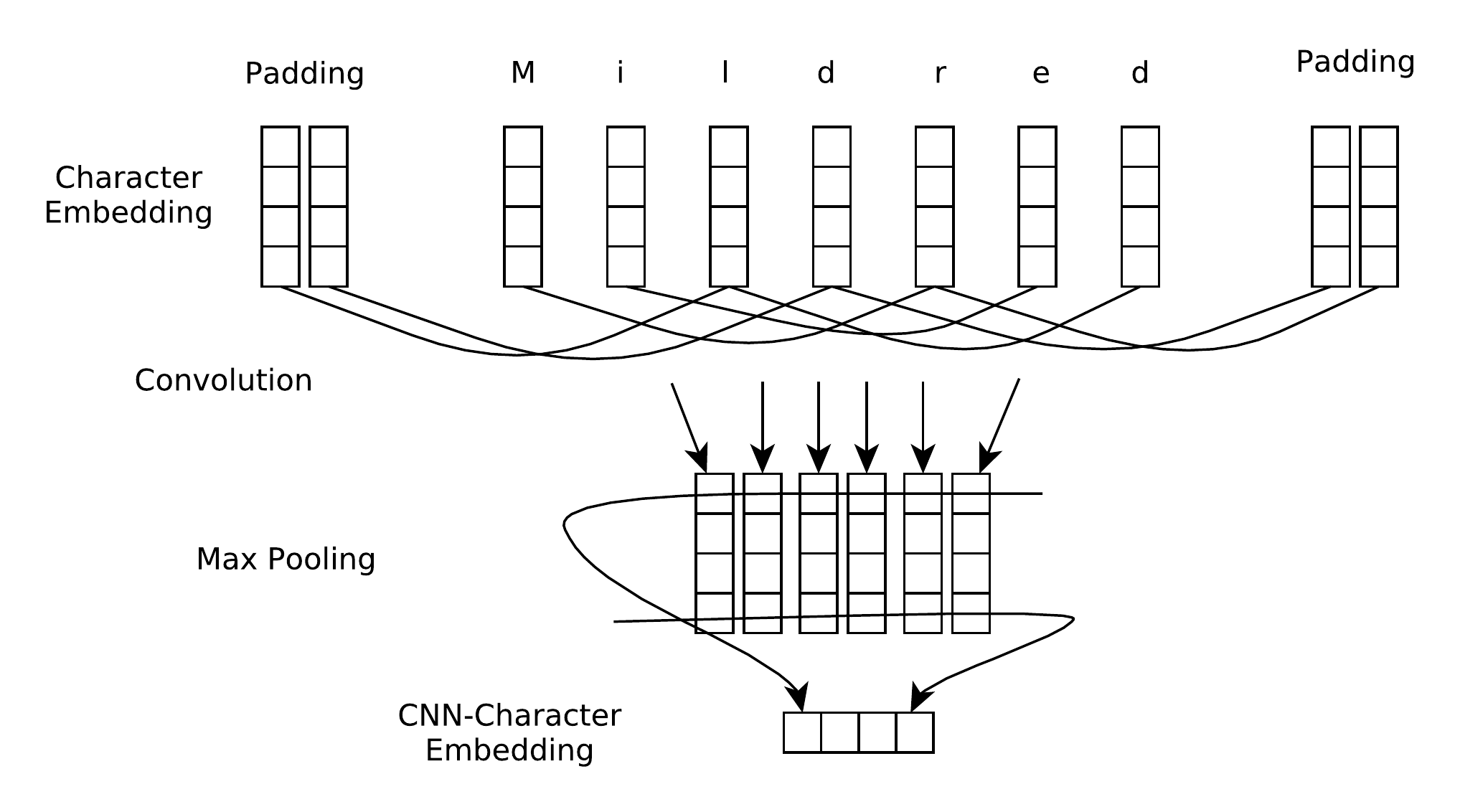}
	\caption{CNN Max Pooling application on Character Embeddings}
	\label{neural:cnn_ner}
\end{figure}

The input sentence is tokenized. Each token in the sentence is converted into a word embedding representation using Glove \cite{glove} vectors (100 dimensional). Each token is also represented in terms of characters by converting the token into a matrix where each vector corresponds to a one-hot encoding vector of a character. The character matrix is fed into a Convolutional Neural Network (CNN) \cite{cnn}. The CNN applies a convolution function to input vectors. We apply a Max-Pooling layer on the CNN output layer that represents the most important character embeddings given the token. This process is shown in Figure~\ref{neural:cnn_ner}. A sigmoid function is applied to the output layer to infer the maximally scoring label for each token.

As the SimpleQuestions dataset does not explicitly provide the subjects, we rely on weak supervision to infer the subject during training process. We infer the position of the subject by querying the inverted index for each n-gram in the question. We assume that the correct subject span is the one that matches the expected subject URI when queried on an index. An algorithm for inferring the span of a subject is given in Algorithm~\ref{alg:ner-train}. These inferred token labels are used as expected output labels from the NER model.

The NER model is trained for 15 epochs, the embedding size of the BiLSTM was 300, the CNN networks uses 3 kernels.

\begin{algorithm}\label{alg:ner}
\caption{Inferring Named Entity Spans}\label{alg:ner-train}
\begin{algorithmic}[1]
\Procedure{Find-Span}{$s,u, m$}\Comment{input sentence \emph{s}, the expected URI \emph{u} and maximum ngram size \emph{m}}
\State $ngrams\gets extract\_ngrams(s, m)$ \Comment{extracts all possible n-grams from the input \emph{s}}
\State $inferred$ = $\emptyset$
\For{each item $n_i$ in $ngrams$}
	\State $candidates$ = retrieve\_candidates($n_i$)
    \If{$u$ is in $candidates$}
		\State $inferred$ = $n_i$
        \State \textbf{Break} \Comment{stops the loop if the expected URI is found}
	\EndIf
\EndFor
\State \textbf{return} $inferred$\Comment{The inferred span for a given URI}
\EndProcedure
\end{algorithmic}
\end{algorithm}

\subsection{Candidate Pair Generation}\label{subsec:neural-candidate-pair}

As shown in Figure~\ref{neural:candidate}, we apply the trained NER system and extract the entity mention, i.e. \emph{Mildred Pierced} in our example.
The extracted mention $m$ is queried on the surface form index. All the matching entries are added to the set $S(m)$. Each entry contains a subject URI (Freebase MID) and a frequency value. For example, the following subjects are found: \emph{m.04t1ftb}, \emph{m.01d13qs}, \emph{m.04t\_038}, \emph{m.0cgv06r}.

We define a $KG$ as a set of triples of the form $(s_i,p_i,o_i)$ that appear in the Freebase-2M dataset. Given a subject $s_i$ we define 
the set $Pred(s_i)$ of all the properties that $s_i$ has as

$$Pred(s_i) := \{p_i \,\, | \,\, \exists o_i (s_i,p_i,o_i) \in KB\}$$

We further define the set of candidate pairs for mention $m$ as:

$$C(m) := \{(s_i,p_i) \,\, | \,\, s_i \in S(m) \wedge p_i \in Pred(s_i)\}$$

For example, the extracted candidate entity \emph{m.01d13qs} has 2 predicates: \emph{music.release\_track.release}, \emph{music.release\_track.recording}. By combining the predicate with the candidate entity we generate candidate pairs (see Figure~\ref{neural:candidate}).

The next step is to find a ranking function that takes an input question text (\emph{q}), the identified mention $m$ and candidate pairs ($C(m)$=\{($s_1$, $p_1$), ($s_2$, $p_2$), ($s_3$, $p_3$), \dots, ($s_n$, $s_n$)\}), and returns the highest ranking pair ($s^*$, $p^*$).

\begin{equation}
\label{candidateEquation}
  ({s}^*,{p}^*) = argmax_{(s_i, p_i) \in C(m)} P(s_i, p_i|q;\theta)
\end{equation}
where $P(s_i, p_i)$ computes the probability of a pair $s_i$ and $p_i$ using the equation below.

\begin{equation}
  P(s_i, p_i|q;\theta) = P(p_i|q;\theta) * P(s_i|q;\theta)
\end{equation}

where $P(p_i|q;\theta)$ is the probability of predicate $p_i$ as computed by our four predicate models described below. $P(s_i|q:\theta)$ is the probability of a subject $s_i$ computed by normalizing the frequency scores retrieved for the mention $m$.

In the following sections, we describe our proposed models for the prediction of target predicates.
\subsection{Model 1: \ModelI}\label{model:1}

Our first model is a BiLSTM classifier that predicts the target predicate given the question text. This is a standard model to predict multiple class labels using a softmax layer by encoding the input text using word and character embeddings. Before passing the question text to our network, we replace the entity name with a special placeholder token \emph{e} (e.g. ``Who wrote e?'') that abstracts away the (inferred) subject mention. Moreover, the model is very similar to the one proposed by \citet{ture2017no}.

\subsubsection{Architecture}
\begin{figure}[ht]	
  \includegraphics[width=0.45\textwidth]{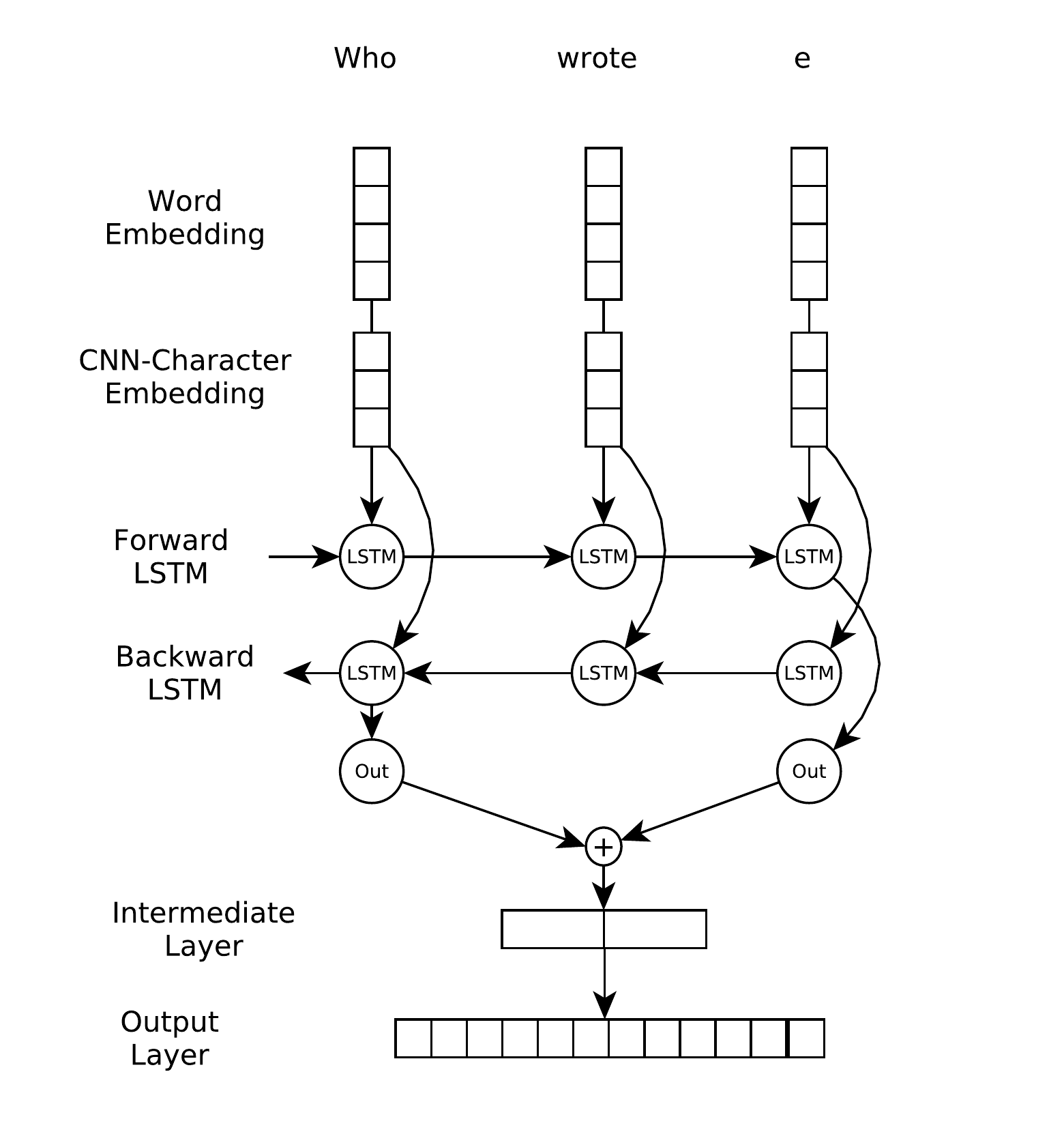}
	\caption{\ModelI model that computes probability distributions for predicates given only the question text}
	\label{neural:model1}
\end{figure}

Similar to the NER model, the question text is encoded on the word and character level. Character-level word embeddings are computed by applying a CNN layer with Max-Pooling on the characters of each token. This process is the same as explained above in Figure~\ref{neural:cnn_ner}. Word and character embeddings are concatenated and passed through a BiLSTM layer. The final states of the BiLSTM are concatenated and fed into a feed-forward layer with softmax activation function, which calculates a probability distribution over a set of predicates. We identified 1629 predicates in the training split of the SimpleQuestions dataset. The architecture is shown in Figure~\ref{neural:model1}. 


The model assigns a probability for each predicate. During candidate pair generation we extract subjects along with frequency values. These frequency values are normalized so that we yield a proper probability for each subject given a question $q$ and a mention $m$. The score for each candidate pair is calculated by multiplying the probability score of the candidate predicate with the normalized frequency value of the candidate subject as given in Equation~\ref{candidateEquation}.

\subsubsection{Hyper-parameters}
The CNN layer uses an embedding size of 100, the LSTM layer uses 200 dimensions; Word embeddings are initialized using 100 dimensional Glove vectors and are retrained with the rest of the model. The model is trained for 100 epochs.

\subsection{Model 2: \ModelII}\label{model:2}
In this subsection, we present a different approach for predicting a predicate from a given question that incorporates pre-trained graph embeddings into the classification process.
Before we describe this model architecture, we first introduce how these graph embeddings are computed.


\subsubsection{Graph Embedding}\label{subsec:neural-graph-embedding} 

There have been different approaches proposed over the years for computing embeddings for knowledge bases. RDF2Vec \cite{rdf2vec} is such a method. By performing a random walk on the graph, the algorithm records paths between pairs of entities. The resulting paths are considered as "sentences" and are fed into the popular word embedding algorithm word2vec which computes vector representations for vertices and edges.

TransE \cite{bordes2013translating} is another method for computing graph embeddings. The method is based on taking a single triple, e.g. $(e_i, p, e_j)$ and creating corrupted triples from it by randomly replacing the subject $e_i$ or the object $e_j$ with a random entity from the KB. The objective of the method is to learn a ranking function that maximizes the margin between the score of an actual triple and the corrupted triples.

In this work, we compute KB embeddings using FastText \cite{fasttext}.
We phrase the task of learning KB embeddings as a classification task. For each triple $t=(e_i, p, e_j)$ in the KB, we construct training samples for the FastText classifier by treating the predicate $p$ and the object $e_j$ as input tokens and subject $e_i$ as the target class. To create embedding vectors that are aware of the role of an entity in a triple, we generate the training sample using role-specific embeddings: $e_i^s$, $e_j^o$ and $p^s$. Here, $e_i^s$ indicates that the target is an entity in the subject position, $e_j^o$ is an input entity in the object position and $p^s$ an input predicate used for predicting a subject entity. 
Analogously, we create a training sample with the object $o$ being the target class.
An example in the FastText format for the triple \texttt{Inferno, hasAuthor, Dan\_Brown} is given below:\\
\begin{itemize}
    \item[] \texttt{\_\_label\_\_Inferno$^s$ hasAuthor$^s$ Dan\_Brown$^o$}
    \item[] \texttt{\_\_label\_\_Dan\_Brown$^o$ hasAuthor$^o$ Inferno$^s$}
\end{itemize}

By training a FastText classifier on the generated training samples, we obtain vector representations for all entities and predicates with respect to their role in the triple\footnote{Due to the huge amount of target classes, training the classifier with a full softmax objective is not feasible. Instead, we use the negative sampling objective that is part of the FastText toolkit as an approximation to the softmax objective.}.
We chose FastText as a classifier for its good performance on text classification tasks in terms of accuracy and speed.


\subsubsection{Architecture}
In the following, we describe a neural network model that uses the pre-trained graph embeddings to predict the target predicate given a question text. The intuition is that we can project the question text into the embedding space of the KG, thus supporting the learning process by utilizing the pre-trained, latent structure of that space. Additionally, the model is not limited to predicates seen during training whereas \ModelI outputs probability distribution to predicates that only appear in the training split.

Similar to the model in Figure~\ref{neural:model1}, the question text is encoded using word and character level embeddings. The encoded text is fed into a BiLSTM layer that outputs a sequence of hidden states. We concatenate the last states of the forward and backward LSTM and pass it through a feed-forward layer which produces a fixed-sized output vector $\hat{p}$ of 200 dimensions. The network is trained to maximize the cosine similarity of the produced output vector $\hat{p}$ and the pre-trained embedding vector $p^*$ of the target predicate.

During prediction we compute the cosine similarity of the computed output vector to the embeddings of all predicates in Freebase-2M and normalize across all predicates to obtain a probability distribution.

The score for a candidate pair is computed as given in Equation~\ref{candidateEquation}.


\begin{figure}[ht]	
  \includegraphics[width=0.5\textwidth]{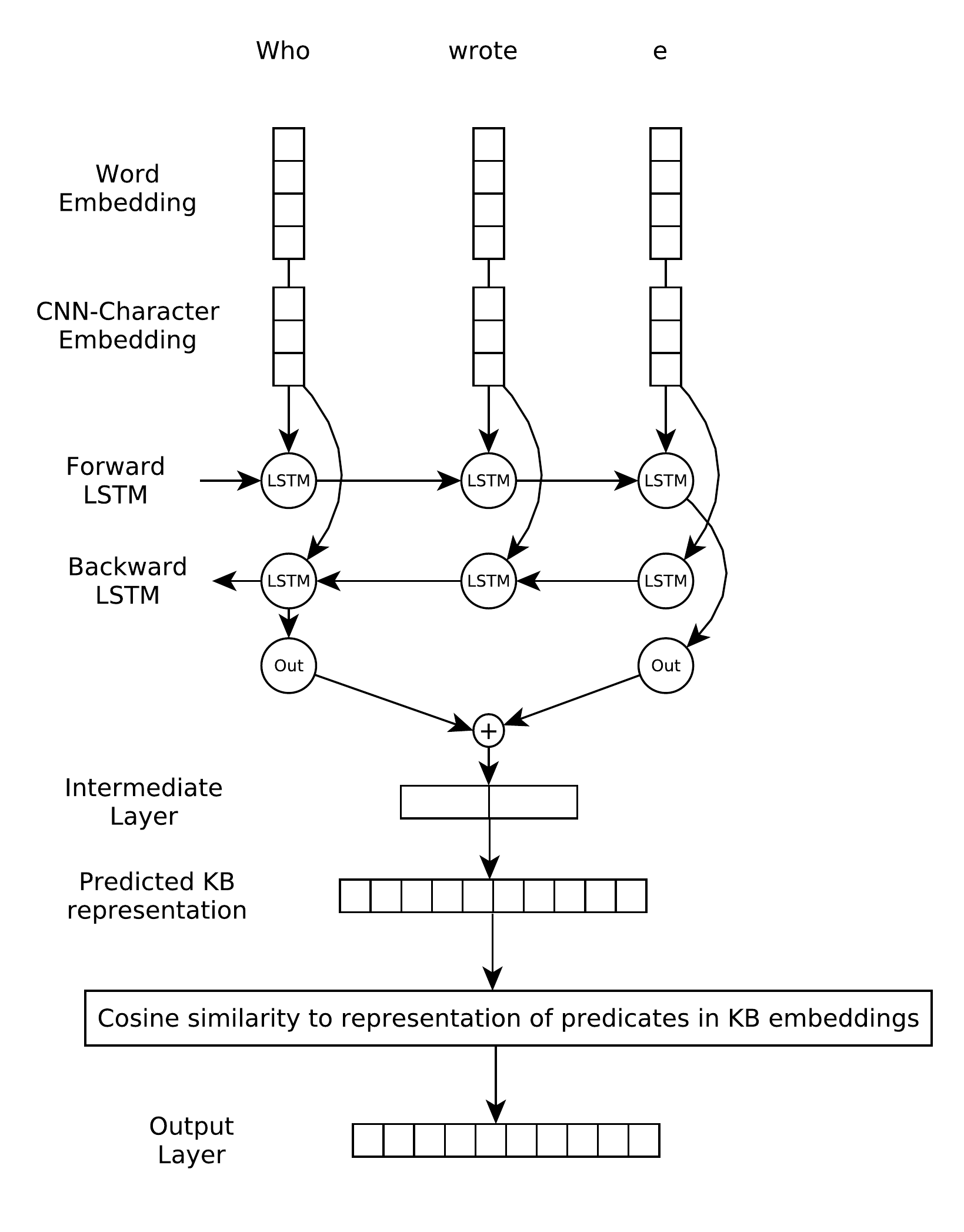}
	\caption{\ModelII model that computes probability distributions for predicates given only the question text}
	\label{neural:model2}
\end{figure}

\subsubsection{Hyper-parameters}
The CNN layer has 100 dimensions, the LSTM has 400 dimensions. We use 100-dimensional Glove vectors.


\subsection{Model 3: \ModelIII}\label{sec:neural-model3}

This model is different than the other 2 models explained above (see Section~\ref{model:1} and Section~\ref{model:2}) in terms of the input to the model. While \ModelII introduces external knowledge about predicates from a knowledge base, this model learns to associate the question text with the tokens in predicate URI. The input is composed of a question text $q$ and the label of a single predicate $p_i$ and the model outputs a binary decision (0 or 1) indicating if the predicate is correct for the question.
By giving the label of a predicate as an input feature, the model can potentially use the similarity between the question text (e.g. \emph{Who \textbf{wrote} e?}) and the predicate label (e.g. \emph{book.\textbf{written}\_work.author}) to determine if the given predicate tokens matches the question text.

\subsubsection{Architecture}
The inputs $q$ and $p_i$ are tokenized and fed into encoding layer that uses word and character embeddings. These are shown as \textbf{BiLSTM Text Encoding}. The encoding is the same process explained in Section~\ref{model:1}(see Figure~\ref{neural:model1}) where the tokens are represented by word and character embeddings and fed into 2-layer BiLSTM.

\begin{figure}[ht]	
  \includegraphics[width=0.45\textwidth]{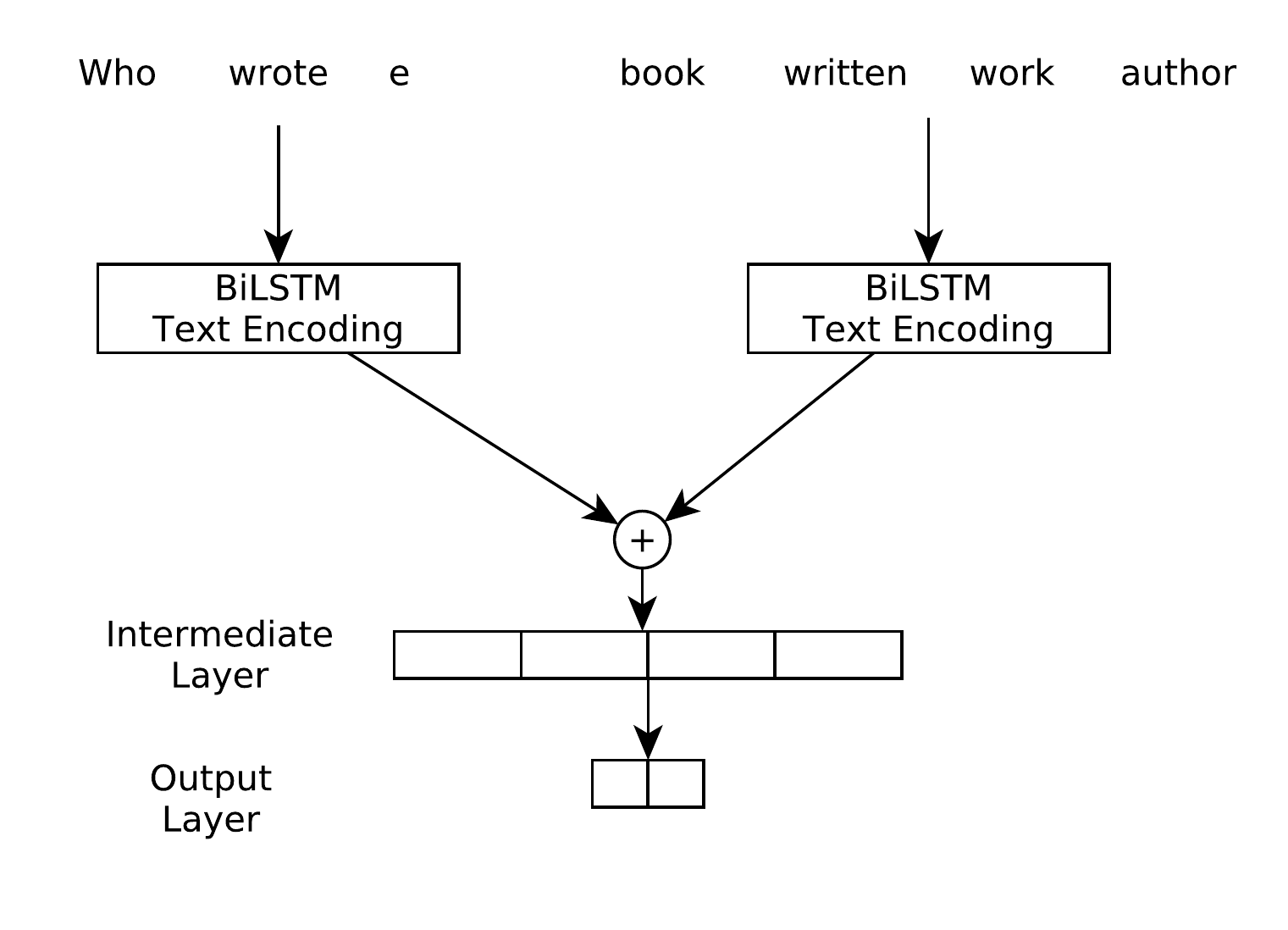}
	\caption{\ModelIII model that computes probability distributions as a binary decision given the question text and the predicate pair}
	\label{neural:model3}
\end{figure}

The tokenization of the predicate $p_i$ is done by splitting the URI by \textit{dot} and \textit{underscore} characters. The latent embeddings are  fed into an intermediate layer, which learns to score the compatibility between (embedded) question input $q$ and predicate $p_i$. Finally, the output layer is a sigmoid function that outputs a binary decision in terms of probability. The model architecture is depicted on Figure~\ref{neural:model3}.

During prediction we collect all predicates from each candidate subject and feed them into the model one at a time. The model outputs a probability for each predicate. The score for a candidate pair is computed using Equation~\ref{candidateEquation}. The highest scoring pair is selected as the final output.

\subsubsection{Hyper-parameters}
We use a CNN with 100 dimensions, and an LSTM with 400 dimensions. We use 100-dimensional Glove vectors. The model is trained for 100 epochs.

\subsection{Model 4: \ModelIV}\label{sec:neural-model4}
For our last model, we train a classifier that predicts the target predicate given the question text using  FastText. The FastText tool implements a linear classifier on top of a bag-of-N-gram representation of a text using word N-grams to preserve local word order and character N-grams for robustness against out-of-vocabulary words. The model outputs a probability for each predicate. The score for a candidate pair is computed using Equation~\ref{candidateEquation}. The highest scoring pair is selected as the final output. For a detailed description of the model architecture we refer to \citet{fasttext}.

\subsubsection{Hyper-parameters}
Due to the moderate size of the target vocabulary\footnote{1629 predicates in the training set.} we can train the classifier with a full softmax objective. We trained the classifier for 50 epochs and a hidden layer size of 100. The classifier uses word N-grams of size 1 and 2 and character N-grams of size 5. 

\section{Evaluation}
\label{evaluation}

We provide evaluations on four models and the building components in isolation as follows:

\begin{enumerate}
\item Named Entity Recognition: the evaluation shows the accuracy for extracting the correct mention from the question text.
\item Named Entity Linking: the evaluation shows in how many cases the subject can be retrieved by index lookup using the detected entity mention from the NER step.
\item Predicate Prediction: this evaluation shows how well the four models perform in predicting the correct predicate for the given question text.
\item Answer Prediction: this evaluation shows how well the proposed models perform on predicting the correct triple and how they compare to other systems on SimpleQuestions dataset.
\end{enumerate}

\subsection{Named Entity Recognition}\label{eva:ner}
\paragraph{Training} We trained a BiLSTM-CRF NER system on SimpleQuestions training split. The model was run for 100 epochs, with word embeddings from Glove 100-dimensional vectors, 200 dimension for LSTM layers.

\paragraph{Prediction} During prediction, we queried all possible n-grams extracted from question text $q$ on a surface form index. N-grams that returned a match were added to a set $N$. A question text $q$ was given as an input to the trained NER system. The output from NER system was compared with each n-gram in $N$. The comparison is based on Edit distance similarity. The N-gram that is the most similar to the output is taken as a recognized subject mention $m$. In this way we can ensure that output mention from NER maps to some set of subjects. The system is regarded as having correctly identified a certain mention $m$ if by looking up the mention in the index the correct subject is returned. For instance, in the Figure~\ref{neural:candidate} the NER system identifies ``mildred pierced'' as a entity mention. By querying the entity mention we retrieve 4 subjects. The expected target subject \emph{m.04t1ftb} is in the list. The NER component achieves an accuracy of 0.82.

\subsection{Named Entity Linking}\label{ev:entityLinking}
Once the subject mention $m$ has been extracted from the NER system, the next step is to get all the matching subjects from the surface form index. We queried the mention $m$ on an index and retrieved subjects with corresponding frequency values.

For evaluation, we ranked the subjects by their frequency values and calculated Recall@K. The system correctly links if the target subject is in the ranked list of K candidate subjects. The results are shown in Table~\ref{table:neural-linking}.

\begin{table}[H]
\caption{Named Entity Linking evaluation on test split using Recall@K} \label{table:neural-linking}
\centering
\begin{tabular}{lc}
\hline
 \bf K & \bf Recall \\
\hline \\[-1ex]
1 & 0.68 \\
2 & 0.74 \\
5 & 0.79 \\
10 & 0.81 \\
25 & 0.82 \\
100 & 0.82 \\
400 & 0.82 \\
\hline
\end{tabular}
\end{table}

\subsection{Predicate Prediction}

All models described above compute probability distributions for predicates. To understand better the building blocks of each model, we evaluated the performance of each model for predicting the correct predicate. Below in Table~\ref{table:neural-predicate-evaluation}, we listed the results for \ModelI, \ModelII,  \ModelIII, and \ModelIV. We trained different models with different hyper-parameters. In Table~\ref{table:neural-predicate-evaluation}, we listed only the best performing models of each type with respective hyper-parameters and their performance scores. The performance score is Accuracy and it was calculated by excluding the subject from the pair and comparing only predicted and expected predicates. As shown in Table~\ref{table:neural-predicate-evaluation}, \ModelIV output performs all other systems while \ModelI and \ModelIII performed similarly.

\begin{table}[H]
\caption{Evaluation of four models on predicate prediction task}
\centering
\begin{tabular}{lc}
\hline
 \bf Name & \bf Accuracy \\
\hline \\[-1ex]
\ModelI & 0.74 \\
\ModelII & 0.68 \\
\ModelIII & 0.73 \\
\ModelIV & \textbf{0.79} \\
\hline
\end{tabular}
 \label{table:neural-predicate-evaluation}
\end{table} 

\subsection{Answer Prediction}
The task of question answering on the SimpleQuestions dataset requires a system to output a single triple consisting of a subject and a predicate. We evaluated the four proposed models on prediction of a triple consisting of a subject and a predicate. The predicated pairs are ranked using Equation~\ref{candidateEquation}.

Moreover, we compared our results with other published systems that evaluated using the same dataset. All results are shown in Table~\ref{table:neural-answer-evaluation}.

\begin{table}[h]
\small
\centering
\caption{Evaluation of four models on answer prediction task}
\begin{tabular}{l c}
\hline 
\bf Name & \bf Accuracy \\ \hline
\ModelI & 0.67\\
\ModelII & 0.61\\
\ModelIII & 0.66\\
\ModelIV & \textbf{0.68}\\
\hline
\end{tabular}
\label{table:neural-answer-evaluation}
\end{table}

\subsection{Error Analysis}

We choose \ModelI to perform error analysis and highlight the errors the model makes. In Table~\ref{table:neural-pair-analysis}, we report the pair prediction results for \ModelI using Recall@K. We extract K top-ranking pairs as given by the model and evaluate how well the system performs on pair prediction. Additionally, we evaluate separately how the subject in the predicted pair compares to the subject of the expected pair. We perform the same evaluate on predicates as well.

We can observe that \ModelI predicts the correct predicate with 0.74 for Recall@1 and 0.8 for Recall@2. The predicate prediction has an upper-bound of 0.84, which was obtained by Recall@20. Subject prediction has a higher performance than pair prediction (0.67 vs 0.74 for Recall@1). Subject prediction has an upper-bound  of 0.82 as explained in the previous section (Section~\ref{ev:entityLinking}). Overall results for pair prediction suggest that the model has the highest margin between Recall@1 and Recall@2. It means that the system could easily reach 0.74 if the ranking function improved.

\begin{table}[H]
\caption{Recall@K values for \ModelI in Pair Prediction task} 
\centering
\begin{tabular}{lccc}
\hline
 K & Pair & Subject & Predicate \\
\hline \\[-1ex]
1 & 0.67 & 0.74 & 0.74\\
2 & 0.74 & 0.78 & 0.80\\
3 & 0.77 & 0.80 & 0.81\\
4 & 0.78 & 0.80 & 0.82\\
5 & 0.79 & 0.81 & 0.83\\
10 & 0.80 & 0.81 & 0.83\\
20 & 0.80 & 0.82 & 0.84\\
\hline
\end{tabular}
\label{table:neural-pair-analysis}
\end{table} 

Next, we analyzed the type of errors systems do as reported in Table~\ref{table:neural-error-analysis}. In total the system predicted 7206 wrong pairs and 14481 correct pairs. In total there are 21687 test instances. We reported the following type of errors:
\begin{itemize}
\item \textbf{Only Wrong Predicate}: If the predicted predicate is incorrect where the predicted subject is correct compared to the target subject and predicate pair.
\item \textbf{Only Wrong Subject}: If the predicted subject is incorrect where the predicted predicate is correct compared to the target subject and predicate pair. These errors could be caused by NER or the frequency value of a subject. 
\item \textbf{Wrong Subject \& Predicate}: If both predicted subject and predicate are incorrect compared to the target subject and predicate pair.
\item \textbf{Empty Prediction}: If both predicate subject and predicate are empty.
\end{itemize}

By picking the highest ranking pair from predictions we compare it to the target pair, if there was a predicted pair. We can see that majority of errors (0.29) are caused by not predicting any pair. The next biggest error mass is in predicting the pair wrong with 0.26. Finally, the system made more errors while predicting the predicate rather than the subject (0.23 vs 0.22).

\begin{table}[H]
\caption{Error analysis for \ModelI in Pair Prediction task} \label{table:neural-error-analysis}
\centering
\begin{tabular}{lcc}
\hline
Error Type & Count & Percentage \\
\hline \\[-1ex]
Only Wrong Predicate & 1642 & 0.23\\
Only Wrong Subject & 1591 & 0.22 \\
Wrong Subject \& Predicate & 1911 & 0.26\\
Empty Prediction & 2062 & 0.29\\
\hline
Total & 7206 & 1.0\\
\hline
\end{tabular}
\end{table}

\subsection{Discussion}

\label{discussion}

We have shown that our NER step is reasonably accurate at detecting the subject span with an accuracy of 0.82. We have seen that in some cases NER picked the wrong span when the question contains some proper name which is not part of a target span, e.g. ``where is mineral hot springs, colorado?'' the expected span is ``mineral hot springs'' while the NER system recognizes the span ``springs, colorado''. Similarly, during entity candidate extraction we have seen that sometimes the target subject has a frequency of 1, which affects the candidate pair score. 

The models \ModelI and \ModelIII performed similarly on predicate prediction while \ModelIII had a margin of 0.6. \ModelIV outperformed all models on predicate prediction. For the answer prediction, \ModelI, \ModelIII and \ModelIV performed similarly even though \ModelIV had the best performance on predicate prediction with a margin more than 0.5.

While none of the model architectures could outperform the current state-of-the-art systems for the overall answer prediction, we evaluated the building blocks of a question answering system and showed how they perform in isolation. It shows how well each component performs and highlights the importance for comparing different models not just on the overall output performance but also the individual small components.

\section{Related Work}

Bordes et al. \citet{simpleQuestions} have presented the first results on the SimpleQuestions dataset. Their approach is based on Memory Networks \citet{memorynetworks}. It generates candidate entities using n-grams from the question text that match some Freebase entity. The approach corrupts the dataset to generate negative samples by assigning random questions from the dataset to Freebase entity and predicate pairs.

Aghaebrahimian et al. \citet{simpleQuestionAgbae} proposed a method for predicting the predicate and subject separately. Their approach uses a 2-layered CNN for ranking predicates. Entity detection is done using the MQL API from Google. Detected entities are disambiguated on the basis of the similarity between the entity's \textit{id} and \textit{name} properties.

Yin et al. \citet{yin2016simple} proposed an approach that uses Convolutional Neural Networks (CNN) with attentive max pooling along with an entity detection and linking system. Their \textbf{active} linking system is based on training a system that learns to detect the span of an entity and retrieves Freebase entities using the mention only from the detected span. The NER system we propose is similar to their approach. They also proposed to use character embeddings in combination with word embeddings since character embeddings generalize better in handling out-of-vocabulary (OOV) words. The overall approach uses character embeddings for encoding entities and word embeddings for predicates. The predicate prediction part in the architecture use a max pooling layer.

Golub et al. \citet{golub2016character} proposed an approach that uses both LSTM and CNN encoders together with character-level embeddings. The question is encoded and fed into a two-layered LSTM with an attention mechanism. Subjects and predicates are also encoded via character-level embeddings and fed into a CNN with two layers. The last layer uses an LSTM layer with attention and outputs a score for a given pair. The authors show that character embeddings generalize better compared to word embeddings (0.78 vs 0.38). The attention layer is shown to be effective, allowing the system to learn to differentiate between entity and predicate spans. We also consider character and words as feature inputs for all our proposed models similar to their approach.

Similarly, Lukovnikov et al. \citet{lukovnikov2017neural} proposed another system that encodes subject and predicate using character and word level embeddings to learn a function that optimizes both subject and predicate assignments by introducing negative samples. Our \ModelIII uses a similar approach by introducing negative samples for predicate prediction.

Ture et al. \citet{ture2017no} proposed a rather simple model based on RNNs without any attention mechanism. They essentially propose to use a model with 2-BiGRU layers for prediction of predicates and a model with 2-BiLSTM layers to predict the span for the subject. Our \ModelI for predicting the property is inspired by their model. However, we could not even come close to reproduce their results with a simplified version of their architecture.


%

\section{Conclusion}
In this paper, we analyze four different model architectures that are evaluated on the SimpleQuestions dataset using the same Named Entity Recognition and Linking system to facilitate the comparison. The results show how well the building components of a QA system perform in isolation and together in a pipeline. 

\ModelIV surprisingly achieves the best performance on predicate and answer prediction where a simple model like FastText performs better than more complex LSTM based models. Additionally, \ModelII introduces external knowledge about predicates in KB but the evaluation results suggest that it does not improve the predicate or the answer prediction.

\section*{Acknowledgements}
This work was supported by the Cluster of Excellence Cognitive Interaction Technology 'CITEC' (EXC 277) at Bielefeld University, which is funded by the German Research Foundation (DFG).

\bibliography{references}
\bibliographystyle{IEEEtranS}

\end{document}